\documentclass[conference]{IEEEtran}
\IEEEoverridecommandlockouts
\usepackage{cite}
\usepackage{amsmath,amssymb,amsfonts}
\usepackage{graphicx}
\usepackage{textcomp}
\usepackage{amsmath}
\usepackage{caption}
\usepackage{subcaption}
\usepackage[ruled,vlined]{algorithm2e}
\usepackage[noend]{algpseudocode}
\usepackage{float}
\usepackage{hyperref}

\newcommand{\pluseq}{\mathrel{+}=}

\def\BibTeX{{\rm B\kern-.05em{\sc i\kern-.025em b}\kern-.08em
    T\kern-.1667em\lower.7ex\hbox{E}\kern-.125emX}}
\begin{document}
\title{DRMIME: Differentiable Mutual Information and Matrix Exponential for Multi-Resolution Image Registration}
\author{Abhishek Nan, Matthew Tennant, Uriel Rubin and Nilanjan Ray
\thanks{This work was supported in part by NSERC Discovery Grants.}
\thanks{Abhisked Nan and Nilanjan Ray are with the Department of Computing Science, Univeristy of Alberta, Edmonton, AB T6G2E8, Canada (e-mails: \{anan1, nray1\}@ualberta.ca). }
\thanks{Matthew Tennant is 
with the Department of Ophthalmology, University of Alberta, Edmonton, Alberta, Canada (e-mail: mtennant@ualberta.ca).}
\thanks{Uriel Rubin is with 
the Department of ophthalmology, Hospital Aleman, Buenos Aires, Argentina (e-mail: urielrubin@gmail.com).}}

\maketitle

\begin{abstract}
In this work, we present a novel unsupervised image registration algorithm. It is differentiable end-to-end and can be used for both multi-modal and mono-modal registration. This is done using mutual information (MI) as a metric. The novelty here is that rather than using traditional ways of approximating MI, we use a neural estimator called MINE and supplement it with matrix exponential for transformation matrix computation. This leads to improved results as compared to the standard algorithms available out-of-the-box in state-of-the-art image registration toolboxes.
\end{abstract}


\section{Introduction}
\label{sec:introduction}

\IEEEPARstart{I}{mage} registration is a common task required for digital imaging related fields that involves aligning two (or more) images of the same objects or scene. In medical image processing, we may wish to perform an analysis of a particular body part over a period of time. Images captured over time, of the same body part or location will change due to changes in the target organ over time as well variability in angle and distance of the target organ from the capture device. The multiple variables over time make image registration an exciting area of research. 

Different imaging modalities can provide different and additive information for the clinician or researcher regarding human tissue. For example, radiation of different wavelengths are able to penetrate human tissues to differing depths. A particular wavelength might be used to produce a map of bone structure, while a different wavelength could be used to map other internal organs. These two different maps are referred to as different modalities. A common way to perform a holistic analysis is to combine the (complimentary) information from these different modalities. Alignment of the different modalities requires multi-modal registration. Metrics that work for mono-modal registration often perform poorly for the multi-modal cases.

One of the most successful metrics used for cross-modal or multi-modal medical image registration is mutual information (MI) \cite{MI_survey}. The most common method used for the computation of MI is histogram-based. As a result, MI suffers from added difficulty of dimensionality when multi-channel (such as color) images are used. Recent MI estimation method such as MINE (mutual information neural estimation) \cite{belghazi2018mine} offers a way to curb this difficulty using a duality principle to estimate a lower bound for MI. Additionally MINE is differentiable because it is computed by neural networks.

Our proposed registration method uses this differentiable mutual information, MINE, so that the automatic differentiation of modern optimization toolboxes, such as PyTorch\cite{NEURIPS2019_9015}, can be utilized. Additionally, our method uses affine transformation computed via matrix exponential of a linear combination of basis matrices. We demonstrate experimentally that transformation matrix computation by matrix exponential yields more accurate registration. Our method also makes use of multi-resolution pyramids. Unlike a conventional method where computation starts at the highest level of the image pyramid and gradually proceeds to the lower levels, we simultaneously use all the levels in gradient descent-based optimization using automatic differentiation.

We refer to our proposed method as DRMIME (differentiable registration with mutual information and matrix exponential). DRMIME is able to achieve state-of-the-art accuracy on two benchmark data sets: FIRE \cite{hernandez2017fire} and ANHIR\cite{ANHIR}.

\section{Background}

\subsection{Optimization for Image Registration}
Let us denote by $T$ the fixed image and by $M$ the moving image to be registered. Let $H$ denote a transformation matrix signifying affine or homography or rigid body or any other suitable transformation. Further, let $Warp(M,H)$ denote a function that transforms the moving image $M$ by the transformation matrix $H.$ Optimization-based image registration minimizes the following objective function to find the optimum transformation matrix $H$ that aligns the transformed moving image with the fixed image:
\begin{equation}
    \min_H D(T,Warp(M,H)),
\label{eqn:opt}
\end{equation}
where $D$ is a loss function that typically measures a distance between the fixed and the warped moving image.

\subsection{Matrix Exponential}

The optimization problem (\ref{eqn:opt}) can be carried out by gradient descent, once we are able to compute the gradient of the loss function $D$ with respect to $H.$ The implicit assumption here is that the loss function $D$ is differentiable and so are the computations within $Warp.$ However, an additional technical difficulty arises in gradient computation when the elements of the transformation matrix $H$ are constrained, as in rigid-body transformation. In such cases, matrix exponential provides a remedy. For example, finding the parameters for rigid transformation can be seen as an optimization problem on a finite dimensional Lie group \cite{schroter2010lie}. In the robotics community, this is a fairly common technique used for the problem of template matching. 

One of the earliest works \cite{taylor1994minimization} shows how to perform optimization procedures over the Lie group \textit{SO(3)} and related manifolds. Their work motivates how any arbitrary geometric transformation has a natural parametrization based on the exponential operator associated with the respective Lie group. They also proved how such a technique is more effective than other methods which approximate gradient descent on the tangent space to the manifold. This was also extended to deformable pattern matching \cite{trouve1998diffeomorphisms}. Among more recent work, data representations in orientation scores, which are functions on the Lie group \textit{SE(2)} were used for template matching\cite{bekkers2017template} via cross-correlation. For brevity, here we just state the mapping for the \textit{Aff(2)} group, which is the group of affine transformations on the 2D plane. This group has 6 generators:\\
\begin{equation*}
\begin{split}
& B_1=
\begin{bmatrix}
0 & 0 & 1\\
0 & 0 & 0\\
0 & 0 & 0\\
\end{bmatrix},
B_2=
\begin{bmatrix}
0 & 0 & 0\\
0 & 0 & 1\\
0 & 0 & 0\\
\end{bmatrix},
B_3=
\begin{bmatrix}
0 & -1 & 0\\
1 & 0 & 0\\
0 & 0 & 0\\
\end{bmatrix}, \\
& B_4=
\begin{bmatrix}
1 & 0 & 0\\
0 & 1 & 0\\
0 & 0 & 0\\
\end{bmatrix},
B_5=
\begin{bmatrix}
1 & 0 & 0\\
0 & -1 & 0\\
0 & 0 & 0\\
\end{bmatrix},
B_6=
\begin{bmatrix}
0 & 1 & 0\\
1 & 0 & 0\\
0 & 0 & 0\\
\end{bmatrix}.
\end{split}
\end{equation*}

If $v=[v_1,v_2,...,v_6]$ is a parameter vector, then the affine transformation matrix is obtained using the expression: $Mexp(\sum_{i=1}^6v_i B_i)$, where $Mexp$ is the matrix exponentiation operation that can be computed by either ($E$ is an identity matrix):\\
\begin{equation}
    Mexp(B) = \lim\limits_{n\to \infty} (E + \frac{1}{n}B)^n,
\end{equation}
or,
\begin{equation}
    Mexp(B) = \sum_{n=0}^{\infty} \frac{B^n}{n!}.    
\label{eqn:mat_exp}
\end{equation}

In DRMIME we use the series (\ref{eqn:mat_exp}) for matrix exponential. We truncate the series after $10$ terms and empirically find that this choice yields good registration accuracy. 

Using the matrix exponential representation for a transformation matrix, the image registration optimization defined in (\ref{eqn:opt}) takes the following form:
\begin{equation}
    \min_{v_1,...,v_6} D(T,Warp(M,Mexp(\sum_{i=1}^6v_i B_i))).
\label{eqn:opt_me}
\end{equation}
We can now apply standard mechanisms of gradient computation $\frac{\partial D}{\partial v_i}$ by automatic differentiation (i.e., chain rule) and adjust parameters $v_i$ by gradient descent.

\subsection{Multi-resolution Computation}

A problem with gradient based methods is that they are highly dependent on initialization and step-size parameters. An alternative approach is to use evolutionary algorithms and/or search heuristics\cite{valsecchi2014intensity}. While both methods have their pros and cons, a lot of modern day machine learning research is focused on developing optimizers for gradient descent and as such is a promising approach. A technique which ameliorates the issues with gradient based methods are multi-resolution pyramids\cite{thevenaz1998pyramid, kruger1998image, alhichri2002multi}. The idea behind the approach is very intuitive; a Gaussian pyramid of images is constructed where the original image lies at the bottom level and subsequent higher levels have a down-scaled, Gaussian blurred version of the image. This not only serves to simplify the optimization, but also serves to speed it up since at the coarsest level the size of the data is greatly reduced making each iteration of gradient descent much faster.

Using a multi-resolution recipe, two image pyramids are built: $T_l$ and $M_l$ for $l=1,...,L,$ where $L$ is the maximum level in the pyramid. Here, $T_1=T$ and $M_1=M$ are the original fixed and moving images, respectively. Then, a registration problem (\ref{eqn:opt_me}) takes the following form:
\begin{equation}
    \min_{v_1,...,v_6} \sum_{l=1}^L D(T_l,Warp(M_l,Mexp(\sum_{i=1}^6v_i B_i))).
\label{eqn:opt_me_mr}
\end{equation}
The usual practice for a multi-resolution approach is to start computation at the highest (i.e., coarsest) level of the pyramid and gradually proceed to the original resolution. In contrast, we found that working simultaneously on all the levels as captured in the optimization problem (\ref{eqn:opt_me_mr}) is more beneficial. 

Note that using the same transformation matrix $Mexp(\sum_{i=1}^6v_i B_i)$ for all resolution levels makes sense only when the image transformation i.e., $Warp$ uses the same canonical range of pixel coordinates at every resolution. For example, our implementation uses the range $[-1,1]\times[-1,1]$ for pixel coordinates. With this view, a multi-resolution pyramid adds more samples in the space $[-1,1]\times[-1,1]$ as we go from lower to higher resolutions.

However, note also that image structures are slightly shifted through multi-resolution image pyramids. So, a transformation matrix suitable for a coarse resolution may need a slight correction when used for a finer resolution. To mitigate this issue, we exploit matrix exponential parameterization and introduce an additional parameter vector $v^1=[v^1_1,...,v^1_6]$ exclusively for the finest resolution level and modify the multi-resolution optimization (\ref{eqn:opt_me_mr}) as follows:
\begin{equation}
    \begin{split}
    \min_{\substack{v_1, \cdots, v_6 \\ v^1_1, \cdots, v^1_6}}
    & \{\sum_{l=2}^L D(T_l,Warp(M_l,Mexp(\sum_{i=1}^6v_i B_i))) + \\
    & D(T_1,Warp(M_1,Mexp(\sum_{i=1}^6(v_i+v^1_i) B_i))) \}.
    \end{split}
\label{eqn:opt_me_mr2}
\end{equation}

\subsection{Metrics for Image Registration}

While there are various metrics used for image registration, probably the simplest is mean squared error (MSE). If successfully registered, the MSE between the fixed and transformed moving image would be close to zero. Often gradient descent based techniques can be used for such intensity-based measures to find the correct registration parameters \cite{klein2009adaptive}. This can also be framed as a supervised learning problem \cite{detone2016deep}, where the goal is to learn the parameters of the homography transformation.

Since different modalities can have different image intensities and varying contrast levels between them, it is unlikely that using MSE as a registration metric will work well. One of the most common metrics used in multi-modality registration is mutual information (MI). MI, in general, is defined as a measure of dependence between two random variables. Two highly dependent variables will have a high MI score, while two less dependent variables will have a low MI score. In the context of image registration, this means that two initially unregistered images will have an MI score which is lower than the MI score between the images once they are completely registered. Gradient-based methods\cite{maes1997multimodality} for MI based image registration work quite well for such cases. In these implementations, MI between two random variables, say $X$ and $Y$ is mathematically quantified by measuring the distance between the joint distribution and the case of complete independence by means of the Kullback-Leibler (KL-) divergence\cite{kullback1997information}:
\begin{equation}
    \label{eqn:MI1}
    MI = \int p_{XY}(x,y) \log \dfrac{p_{XY}(x,y)}{p_X(x)p_Y(y)}dxdy,
\end{equation}
where $P_{XZ}$ is the joint density for random variables $X$ and $Z.$ $P_{X}$ and $P_{Z}$ are marginal densities for $X$ and $Z,$ respectively. 
Again, in the context of scalar-valued images, these joint probabilities are calculated using a two-dimensional histogram of the two images. Most current MI-based techniques for registration use slight variations of the above method to approximate MI. While this works well, there are some issues associated with this method of evaluation as follows.
\begin{itemize}
    \item The number of histogram bins chosen becomes a hyperparamter. While increasing the number of bins would lead to better accuracy in computation, this comes at the cost of time. Furthermore, there is no theoretical upper bound on the number of bins that should be used for accurate results.
    \item Images with higher dimensions (color images, hyper-spectral images), would need a higher dimensional histograms and joint a histogram requiring a very large sample that is often computationally prohibitive. For instance, an RGB image has 3 channels and that would need a 6-dimensional joint histogram. A common way to bypass this restriction is to work with grayscale intensities of images, but this leads to loss of valuable information, incorporating which would very likely have led to better results.
\end{itemize}

A potential solution to the above problem is presented by MINE \cite{belghazi2018mine} that uses the Donsker-Varadhan (DV) duality to compute MI (we provide a simple proof at the Appendix):
\begin{equation}
    MI = \sup_{f} J(f),
\label{eqn:MI2}
\end{equation}
where $J(f)$ is the DV lower bound:
\begin{equation}
\begin{split}
    J(f) = & \int f(x,z)P_{XZ}(x,z)dxdz - \\
    log( & \int exp(f(x,z)) P_{X}(x)P_{Z}(z)dxdz).
\end{split}
\label{eqn:DV}
\end{equation}
MINE uses a neural network to compute $f(x,z)$ and uses Monte Carlo technique to approximate the right hand side of (\ref{eqn:DV}). MINE claims that computations of (\ref{eqn:MI2}) scales much better than histogram-based computation of MI \cite{belghazi2018mine}. 

The optimization for image registration (\ref{eqn:opt_me_mr2}) using mutual information now becomes:
\begin{equation}
    \begin{split}
    \max_{\substack{v_1, \cdots, v_6 \\ v^1_1, \cdots, v^1_6 \\ \theta}}
    & \{\sum_{l=2}^L MINE(T_l,Warp(M_l,Mexp(\sum_{i=1}^6v_i B_i))) + \\
    & MINE(T_1,Warp(M_1,Mexp(\sum_{i=1}^6(v_i+v^1_i) B_i))) \},
    \end{split}
\label{eqn:opt_drmime}
\end{equation}
where $\theta$ denotes the parameters of the neural network that MINE uses to realize $f.$ Notation $MINE(P,Q)$ in (\ref{eqn:opt_drmime}) is used to denote DV lower bound (\ref{eqn:DV}) computed on two images $P$ and $Q$. Since DV lower bound is differentiable because a neural network (henceforth referred to as MINEnet) realizes the function $f,$ we can use automatic differentiation for gradient ascent optimization (\ref{eqn:opt_drmime}).

\section{DRMIME Algorithm}
\begin{figure*}[h]
    \centering
    \includegraphics[scale=0.55]{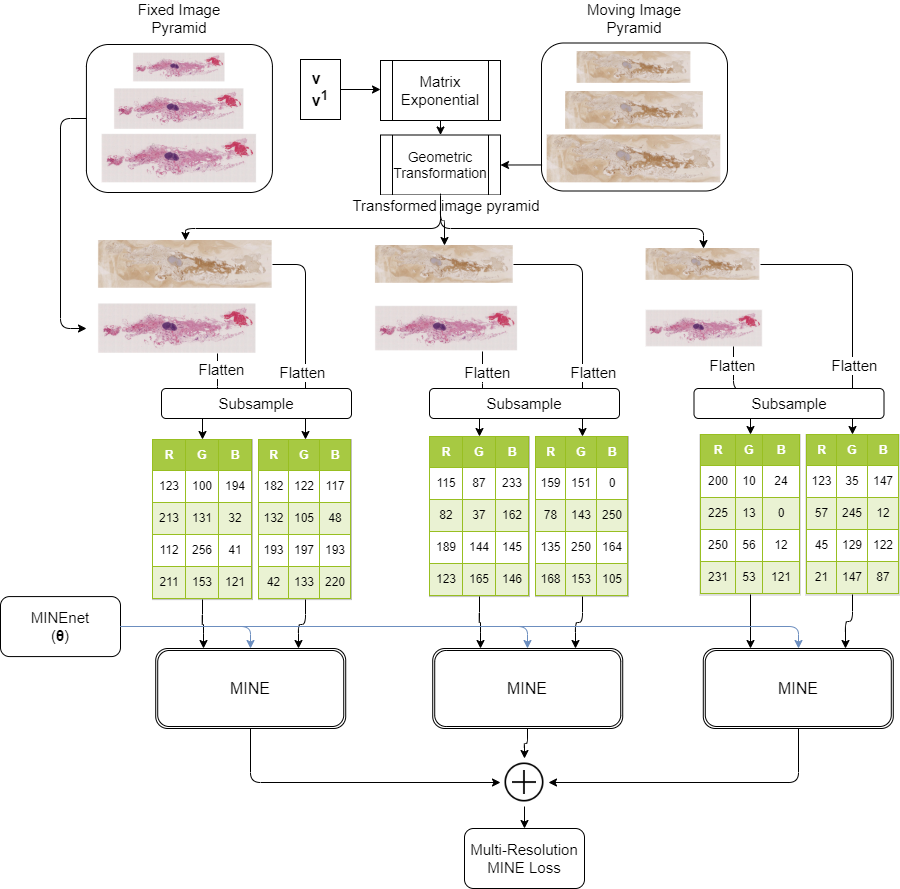}
    \caption{Pipeline for the DRMIME Registration algorithm}
    \label{fig:full_algorithm}
\end{figure*}

Our proposed image registration method DRMIME is illustrated in Fig. \ref{fig:full_algorithm} that implements the optimization problem (\ref{eqn:opt_drmime}).  Algorithm \ref{alg:DRMIME} implements DRMIME that uses DV lower bound (\ref{eqn:DV}) MINE computed in turn by Algorithm \ref{alg:MINE}, which employs a fully connected neural network MINEnet. MINEnet has two hidden layers with $100$ neurons in each layer. We use ReLU non-linearity in both the hidden layers. Appendix contains details about implementation including learning rates, hyperparameters and optimizations used. The code for DRMIME is available on \href{https://github.com/abnan/DRMIME}{GitHub}.

Algorithm \ref{alg:MINE} takes in two images $P$ and $Q$ along with a subset of pixel locations $I.$ It creates a random permutation $I^s$ of the indices $I.$ $I_i$ denotes the $i^{\text{th}}$ entry in the index list $I$, while $P_{I_i}$ deontes the $I_i^{\text{th}}$ pixel location on image $P.$

Algorithm \ref{alg:DRMIME} starts off by building two image pyramids, one for the fixed and another for the moving image. Due to memory constraints, especially for GPU, a few pixel locations are sampled that enter actual computations. This step appears as ``Subsample'' in Fig. \ref{fig:full_algorithm}. We have used two variations of sampling: (a) randomly choosing only 10\% of pixels locations in each iteration and (b) finding Canny edges\cite{canny1986computational} on the fixed image and choosing only the edge pixels. Results section shows a comparison between these two options. Fig. \ref{fig:full_algorithm} illustrates two other computation modules - ``Matrix Exponential'' and ``Geometric Transformation'' that denotes $Mexp$ and $Warp$ operations, respectively. 

\begin{algorithm}[h]
\SetAlgoLined
%\State $\theta \gets$ initialize MINEnet parameters

\textbf{Input:} Image $P$, Image $Q$, Sampled pixel locations $I$ \\
\textbf{Output:} Estimated mutual information (a real number) \\
\textbf{MINE($P, Q, I$)}

\Indp
    Shuffle indices: $I^s = RandomPermute(I)$ \;
    $N = length(I)$ \;
    Return Donsker-Varadhan lower bound \cite{belghazi2018mine}: $\frac{1}{N}\sum_{i}MINEnet_\theta(P_{I_i}, Q_{I_i}) - log(\frac{1}{N}\sum_{i}exp(MINEnet_\theta(P_{I_i}, Q_{I^s_i})))$ \;
\caption{MINE}
\label{alg:MINE}
\end{algorithm}

\begin{algorithm}[h]
\SetAlgoLined
    Build multiresolution image pyramids $\{T_l,M_l\}_{l=1}^{L}$ \;
    Set learning rates $\alpha$, $\beta$ and $\gamma$\;
    Use random initialization for MINEnet parameters $\theta$ \;
    Initialize $v$ and $v^1$ to the 0 vectors \;
    \For {each iteration}{
        $MI = 0$ \;
        $H_1 = Mexp(\sum_{i=1}^6 (v_i+v^1_i) B_i)$ \;
        $I_1 = \text{Sample pixel locations on } T_1$ \;
        $ MI \pluseq MINE(T_1, Warp(M_1, H_1), I_1)$ \; 
        $H = Mexp(\sum_{i=1}^6 v_i B_i)$ \;
        \For {$l = [2,L]$}{
            $I_l = \text{Sample pixel locations on } T_l$ \;
            $ MI \pluseq MINE(T_l, Warp(M_l, H), I_l)$ \;
        }
        Update MINEnet parameter: $\theta \pluseq \alpha \nabla_{\theta} MI$ \;
        Update matrix exponential parameters: $v \pluseq \beta \nabla_v MI$ and $v^1 \pluseq \gamma \nabla_{v^1} MI$\;
    }
    Compute final transformation matrix: $H_1 = Mexp(\sum_{i=1}^6 (v_i+v^1_i) B_i)$ \;
\caption{DRMIME}
\label{alg:DRMIME}
\end{algorithm}

\section{Datasets}
The datasets chosen for our experiments correspond to testing two important hypotheses. First, performing image registration with our algorithm on images within the same modality fares comparably (or better) to other standard algorithms. For this, we use the FIRE dataset \cite{hernandez2017fire}. Second, since our algorithm is based on MI, it can handle multi-modal registration successfully as well. For this we use data from the ANHIR (Automatic Non-rigid Histological Image Registration) 2019 challenge\cite{ANHIR}. Note that ANHIR contains color images that further tests the capability of DRMIME to handle multi-channel images.

\subsection{FIRE}
The FIRE dataset provides 134 retinal fundus image pairs divided into 3 categories: S(71 pairs), P(49 pairs) and A(14 pairs). The primary uses of the categories being Super Resolution, Mosaicing and Longitudinal Study, respectively. While categories S and A have $>75\%$ overlap, category P has very little overlap ($<75\%$) and so none of the algorithms we evaluated (including ours) perform well on P category, leading to little or no registration in most cases (even diverging in some instances). So for a fair evaluation, we leave out category P.

\subsubsection{Ground Truth}
The FIRE dataset provides the location of 10 points in each image and the location of the corresponding 10 points in the paired (to-be-registered) image. These points were obtained by annotations created by experts and further refined to mitigate human error \cite{hernandez2017fire}.
\subsubsection{Evaluation}
For a perfectly registered pair of images, the points from each image will completely overlap; this means that the average euclidean distance (AED) between the points after registration should be close/equal to 0. We calculate the AED between these points as a measure of the registration accuracy of each algorithm. We also use normalized co-ordinates (image co-ordinates vary between 0 and 1) to calculate the AED so that we can have a uniform scale for all images irrespective of the size of the images. We call this metric Normalized Average Euclidean Distance (NAED).
\subsubsection{Preprocessing}
Each image in this dataset is $2912 \times 2912$ pixels, but only the central portion of the images contain the retinal fundus, the rest of the image being black. While it's possible to use masks to remedy this, not all frameworks support masks, so in order to have a fair comparison across all algorithms, we crop these images to include only the retinal fundus. The cropping was selected such that it includes no blank (black) space and it remains rectangular (square). The cropped area was $1941 \times 1941$ pixels.

\subsection{ANHIR}
The ANHIR dataset provides pairs of 2D microscopy images of histopathology tissue samples stained with different dyes\cite{ANHIR}. The task is difficult due to non-linear deformations affecting the tissue samples, different appearance of each stain, repetitive texture, and the large size of the whole slide images.
\subsubsection{Ground Truth}
This dataset provides the ground truth in a format similar to the FIRE dataset, with the exception being each image pair usually has more than 10 corresponding landmark points.
\subsubsection{Evaluation}
Only 230 pairs are available with their ground truth as part of the training data, so we only evaluate on this set of images. We report NAED after the registration process (same as the FIRE dataset).
\subsubsection{Preprocessing}
The ANHIR dataset has extremely high resolution pictures (some categories go upto $65k \times 60k$ pixels on average) and some registration frameworks fail to process such large images. Furthermore, different stainings of the same tissue have different resolutions as well. To solve these two problems when registering a pair of images, they are scaled down by a factor of 5 while keeping the original aspect ratio; this solves the first problem. Then the image with the smaller aspect ratio is rescaled to match the width of the image with the larger aspect ratio and the top and bottom of the smaller one are padded to match the height of the larger. This way we keep the aspect ratio of the original images with no distortions and still arrive at a common and smaller, more manageable resolution.

\begin{figure*}[h!]
    \centering
    \includegraphics[scale=0.45]{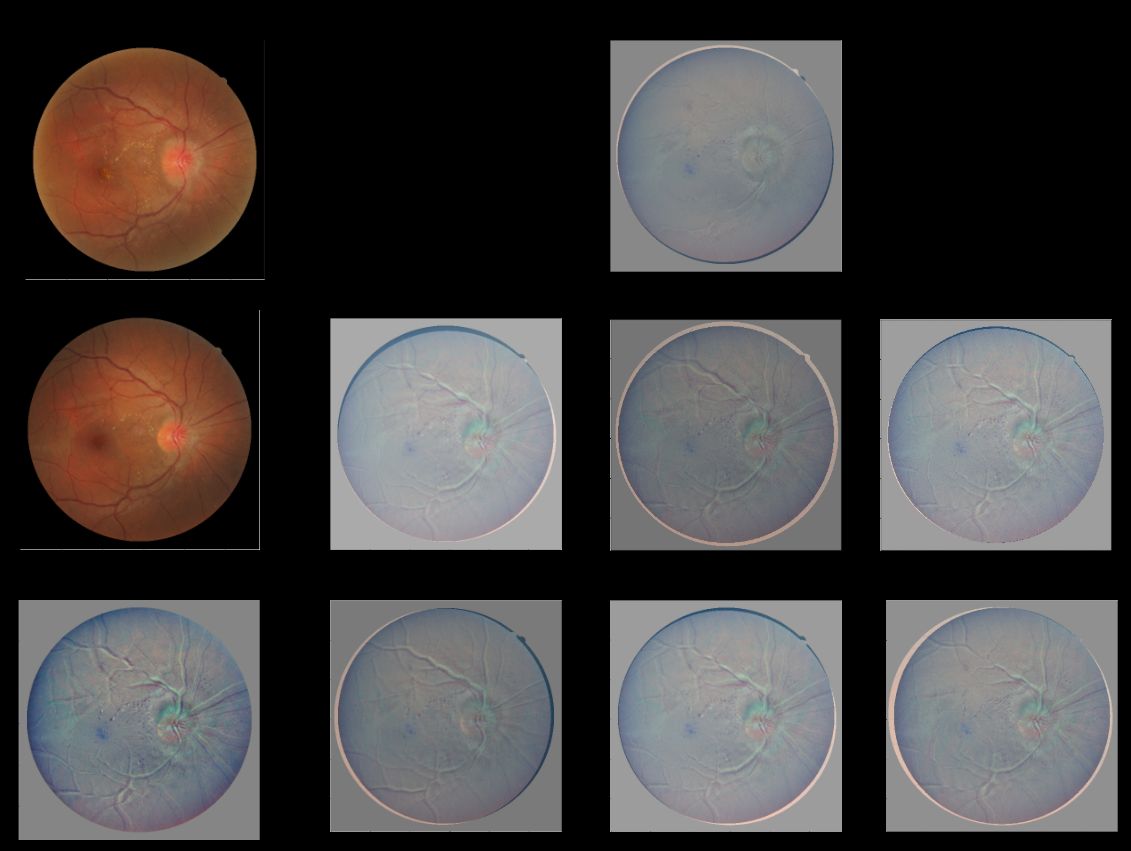}
    \caption{The images on the left show a pair to be registered from the FIRE dataset. The images on the right represent the difference between the transformed moving image and the fixed image after registration by different algorithms.}
    \label{fig:fire_samples}
\end{figure*}

\section{Experiments}
We evaluate our method against the following off-the-shelf registration algorithms from popular registration frameworks. The competing algorithms were whether they use MI or can be used for multi-modal registration:
\begin{enumerate}
    \item\textbf{Mattes Mutual Information (MMI)} \cite{mattes2001nonrigid, mattes2003pet, MMI}: As mentioned in equation (\ref{eqn:MI1}), we need to compute the joint and marginal probabilities of the fixed and moving images. To reduce the effects of quantization from interpolation and discretization due to binning, this version of MI computation uses Parzen windowing to form continuous estimates of the underlying image histogram.
    
    \item \textbf{Joint Histogram Mutual Information (JHMI)} \cite{thevenaz2000optimization, JHMI}: This method computes Mutual Information using Parzen windows as well, but it uses separable Parzen windows. By selection of a Parzen window that satisfies the partition of unity, it provides a tractable closed-form expression of the gradient of the MI computation with respect to transformation parameters.
    
    \item \textbf{Normalized Cross Correlation (NCC)}\cite{NCC}: As the names says, the correlation between the moving and the fixed image pixel intensities is computed. The correlation is normalized by the autocorrelations of both the fixed and moving images.
    
    \item \textbf{Mean Square Error (MSE)}\cite{MSE}: This is the mean squared difference of the pixelwise intensity between the fixed and moving image.
    
    \item \textbf{AirLab Mutual Information (AMI)}\cite{DBLP:journals/corr/abs-1806-09907}: AirLab is a PyTorch based image registration framework. It performs histogram based mutual information computation\cite{viola1997alignment,maes1997multimodality}. Since it is a deep learning based solution, it provides support for using batches as well as state-of-the-art optimizers and GPU support.
    
    \item \textbf{Normalized Mutual Information (NMI)}\cite{studholme1999overlap, NMI}: The initial PDF (probability density function) construction is done using Parzen histograms, and then MI is obtained by double summing over the discrete PDF values. In this metric, the final MI is normalized to a range between 0 and 1.
\end{enumerate}

Also as a note, most libraries limit 2D image registration to affine transforms in terms of degrees of freedom. While it is possible to use perspective transforms with our algorithm just by changing the base vector ($v$), in order to have a fair comparison, we limit our algorithm to affine transforms as well.
The implementations of the above algorithms were used from these packages:
\begin{itemize}
    \item SITK: MMI, JHMI, NCC, MSE
    \item AirLab: AMI
    \item SimpleElastix: NMI
\end{itemize}

\section{Results}
This section lists the results for all algorithms on the aforementioned datasets. For all evaluations, we also conduct a paired t-test with DRMIME to investigate if the results are statistically significant (p-value $<$ 0.05).

Fig. \ref{fig:fire_samples} shows registration results for a random sample. Table \ref{tab:fire_res} shows the NAED for all algorithms on the FIRE dataset. Here, DRMIME performs almost an order of magnitude better than the competing algorithms and the results are statistically significant.
    \begin{table}[H]
    \caption{NAED for FIRE dataset along with paired t-test significance values}
    \centering
    \begin{tabular}{||c|c|c||}
        \hline
        Algorithm & NAED (Mean $\pm$ STD) & p-value \\ 
         \hline\hline
         DRMIME & \textbf{0.0048} $\pm$ 0.014 & - \\ 
         \hline
         NCC & 0.0194 $\pm$ 0.033 & 1.3e-04 \\
         \hline
         MMI & 0.0198 $\pm$ 0.034 & 5.4e-05 \\
         \hline
         NMI & 0.0228 $\pm$ 0.032 & 1.7e-08 \\
         \hline
         JHMI & 0.0311 $\pm$ 0.046 & 4.5e-07 \\
         \hline
         AMI & 0.0441 $\pm$ 0.028 & 1.4e-27 \\
         \hline
         MSE & 0.0641 $\pm$ 0.094 & 3.5e-03 \\ [1ex] 
        \hline
    \end{tabular}
    \label{tab:fire_res}
    \end{table}

\begin{figure}[H]
\centering
\includegraphics[scale=0.35]{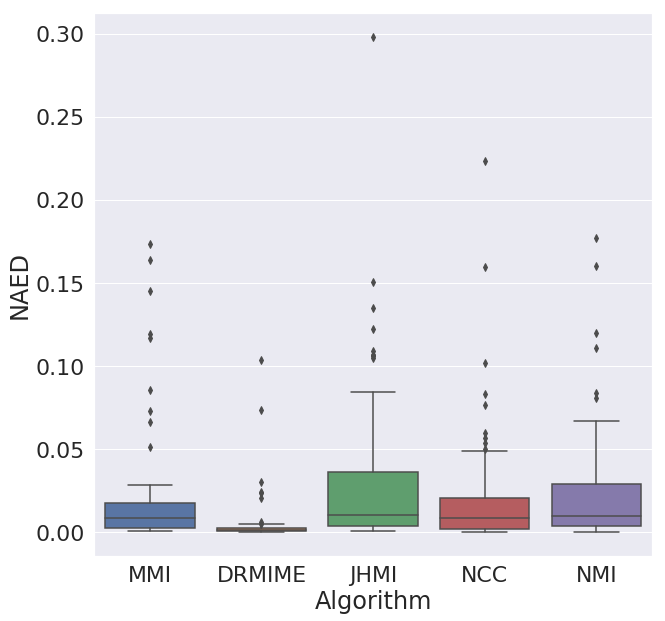}
\caption{Box plot for NAED of the best 5 performing algorithms on FIRE}
\label{fig:fire_res}
\end{figure}

Fig. \ref{fig:fire_res} presents a closer look at the same metrics from Table \ref{tab:fire_res}. We note that DRMIME has very few outliers due to the robustness of the algorithm.

\begin{figure*}[h!]
    \centering
    \includegraphics[scale=0.4]{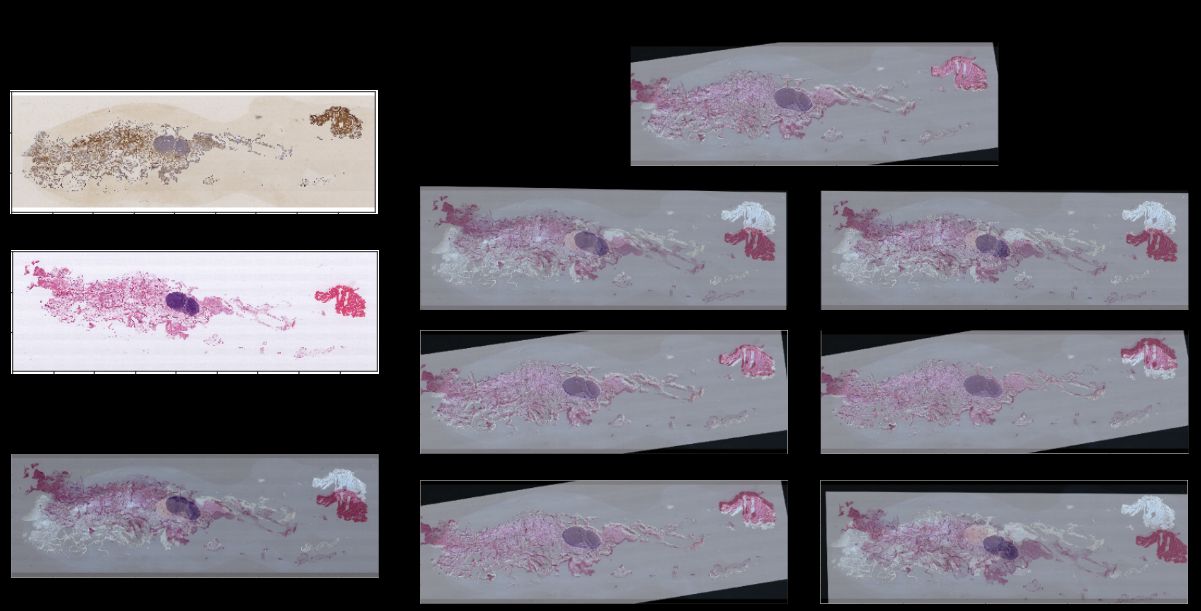}
    \caption{The images on the left show a pair to be registered from the ANHIR dataset. The images on the right represent the difference between the transformed moving image and the fixed image after registration by different algorithms.}
    \label{fig:anhir_samples}
\end{figure*}
Fig. \ref{fig:anhir_samples} shows registration results for a random sample. Table \ref{tab:anhir_res} presents the NAED metrics for the ANHIR dataset. While the margin of improvement is not as large as in case of the FIRE dataset, DRMIME is still statistically the best performing algorithm.

    \begin{table}[H]
    \caption{NAED for ANHIR dataset along with paired t-test significance values}
    \centering
    \begin{tabular}{||c|c|c||}
        \hline
        Algorithm & NAED (Mean $\pm$ STD) & p-value \\ 
         \hline\hline
         DRMIME & \textbf{0.0384} $\pm$ 0.087 & - \\ 
         \hline
         NCC & 0.0461 $\pm$ 0.084 & 7.0e-04 \\
         \hline
         MMI & 0.0490 $\pm$ 0.082 & 6.2e-05 \\
         \hline
         MSE & 0.0641 $\pm$ 0.094 & 5.5e-14 \\
         \hline
         NMI & 0.0765 $\pm$ 0.090 & 3.0e-31 \\
        \hline
         AMI & 0.0769 $\pm$ 0.090 & 3.7e-30 \\
         \hline
        JHMI & 0.0827 $\pm$ 0.100 & 8.3e-21 \\  [1ex] 
         \hline
    \end{tabular}
    \label{tab:anhir_res}
    \end{table}

\begin{figure}[H]
\centering
\includegraphics[scale=0.35]{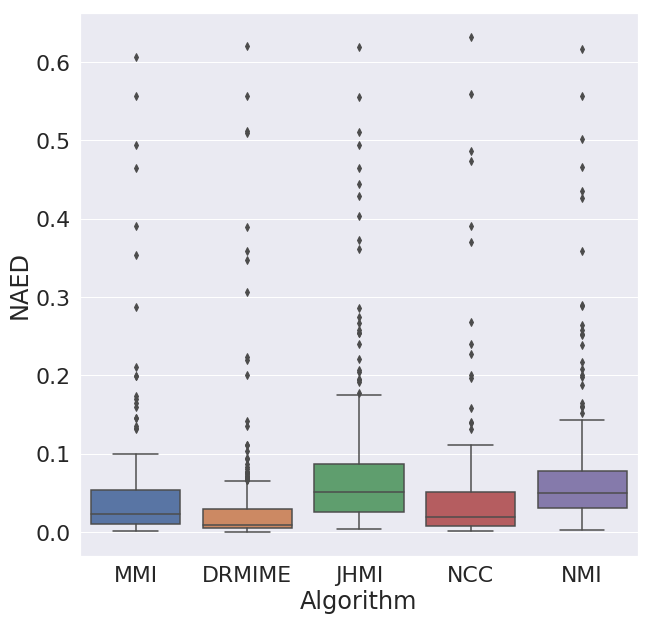}
\caption{Box plot for top 5 performing algorithms on ANHIR}
\label{fig:anhir_res}
\end{figure}

The box-plots in Fig. \ref{fig:anhir_res} also emphasise the same conclusion as we saw before, i.e. DRMIME outperforms the other competing algorithms.

\section{Efficiency}
For efficiency we look at two perspectives: time efficiency and accuracy. On a set of 10 randomly selected images (the set remains the same across all algorithms) from the FIRE dataset, we run these two sets of experiments for all the algorithms. We report the registration accuracy in terms of the ground truth (NAED) of these 10 images. The hardware for these experiments was NVIDIA GeForce GTX 1080 Ti, Intel(R) Xeon(R) CPU E5-2620 v4 @ 2.10GHz, 32GB RAM.

For time efficiency, we run each algorithm for 1000 epochs, report the time taken and the accuracy achieved. The time taken tells us the fastest algorithm among those being considered, and also at the same time, its accuracy should at least be on par with other slower algorithms.
% 22671.17363 sec
\begin{center}
    \begin{table}[!h]
    \caption{Time taken for 1000 epochs and resultant NAED (lower is better)}
    \centering
    \begin{tabular}{||c|c c||}
        \hline
        Algorithm & Time (seconds) & NAED \\ 
         \hline\hline
         DRMIME (50 epochs) & \textbf{58} & 0.02037\\
         \hline
         NMI & 60 & 0.02503\\
         \hline
         AMI & 620 & 0.02942\\ 
         \hline
         DRMIME & 1425 & \textbf{0.00368} \\ 
         \hline
         MMI & 2904 & 0.00598\\
         \hline
         JHMI & 1859 & 0.00605\\
         \hline
         NCC & 3804 & 0.00697\\
          \hline
         MSE & 2847 & 0.02918\\ [1ex] 
        \hline
    \end{tabular}
    \label{tab:perf_table1}
\end{table}
\end{center}

From Table \ref{tab:perf_table1}, we can infer that while our algorithm attains the best NAED, it ranks third in terms of time taken to execute 1000 epochs. While AMI and NMI are faster, they are almost an order of magnitude worse in terms of the NAED performance.

Since this a tradeoff between time and efficiency, DRMIME can perform extremely well at both ends of the spectrum. For instance, while individual epochs on AMI and NMI might be faster, we can achieve comparable accuracy by running DRMIME for much less epochs; within 50 epochs of optimization DRMIME achieves an NAED of 0.02037 taking only 58 seconds. The reason for a single epoch taking longer for DRMIME can be attributed to the fact that it works with batched data.

Also as a note, only DRMIME and AMI are GPU compatible, while the remaining were run on CPU.

\section{Ablation study}
In this section we perform several ablation studies to have an understanding of the roles of all the components used in DRMIME, such as multi-resolution pyramids, matrix exponential and smart feature selection via Canny edge detection. We compare the performance of DRMIME to versions of it without using the aforementioned components.

\subsection{Effect of multi-resolution}
All hyperparameters are kept the same in the with and without experiments, the only difference being in the with multi-resolution experiment we use 6 levels of the Gaussian pyramids in the DRMIME algorithm, whereas in the without experiment we have a single level which is the native resolution of the image. Table \ref{tab:ablation_table1} lists the results for these experiments.

\begin{center}
    \begin{table}[!h]
    \caption{NAED for MINE with and without using multi-resolution pyramids}
    \centering
    \begin{tabular}{||c|c|c|c||}
        \hline
        \textbf{Dataset} & \textbf{DRMIME} & \textbf{Without MultiRes} & \textbf{p-value} \\ 
         \hline\hline
         FIRE & 0.0048 $\pm$ 0.014 & \textbf{0.0043} $\pm$ 0.014 & 0.365 \\ 
         \hline
         ANHIR & \textbf{0.0384} $\pm$ 0.087 & 0.1089 $\pm$ 0.150 & 1.78e-15\\
         \hline
    \end{tabular}
    \label{tab:ablation_table1}
\end{table}
\end{center}

While the idea of multi-resolution was introduced in image registration to facilitate optimization, we note that many of the off-the-shelf algorithms have the same learning rate for all levels. As we are working with only an approximation of the distribution of the original data at different levels of the pyramid, there is a small chance that optimization at a particular sublevel could diverge. This leads to poor registration results occasionally. In our implementation of DRMIME, we produce batches which include data from all levels of the pyramid, making the optimization process much more robust, faster and less prone to divergence. Fig. \ref{fig:fire_res} provides evidence to this since very few results fall outside the interquartile range (as comapared to other algorithms).

\subsection{Effect of matrix exponentiation}
All hyperparameters are again kept the same in the with and without experiments; the only difference being, that rather than using a manifold basis vector, we now have 6 parameters indicating the degrees of freedom of an affine transform in a transformation matrix, i.e.
\begin{equation*}
\begin{bmatrix}
\theta_1 & \theta_2 & \theta_3\\
\theta_4 & \theta_5 & \theta_6\\
0 & 0 & 1\\
\end{bmatrix}.
\end{equation*}

\begin{center}
    \begin{table}[!h]
    \caption{NAED for MINE with and without using manifolds}
    \label{tab:ablation_table2}
    \centering
    \begin{tabular}{||c|c|c|c||}
        \hline
        \textbf{Dataset} & \textbf{DRMIME} & \textbf{Without Manifolds} & \textbf{p-value} \\ 
         \hline\hline
         FIRE & 0.0048 $\pm$ 0.014 & \textbf{0.0045} $\pm$ 0.015 & 0.4933 \\ 
         \hline
         ANHIR & \textbf{0.0384} $\pm$ 0.087 & 0.0580 $\pm$ 0.134 & 0.0012\\
         \hline
    \end{tabular}
\end{table}
\end{center}

Table \ref{tab:ablation_table2} presents the results for these experiments. While the ablation study on the FIRE dataset results in similar results, the p-values from the paired t-test tells us that the results are not very significant to be able to interpret anything. The ANHIR datset on the other hand sees a statistically significant improvement with use of matrix exponentiation.

\subsection{Effect of Sampling strategy}
It could be argued that our smart feature extraction via Canny edge detection helps DRMIME perform better than other algorithms, since other algorithms do not have such custom feature detectors embedded in their pipeline. In order to reduce this potential confounding variable, we also assessed the performance of DRMIME with random sampling as well to make a fair comparison.

\begin{center}
    \begin{table}[!h]
    \caption{NAED for MINE with Canny edge detection and Random Sampling (10\%)}
    \centering
    \begin{tabular}{||c|c|c|c||}
        \hline
        \textbf{Dataset} & \textbf{With Canny} & \textbf{Random Sampling(10\%)} & \textbf{p-value} \\ 
         \hline\hline
         FIRE & \textbf{0.0048} $\pm$ 0.014 & 0.0097 $\pm$ 0.026 & 0.0296 \\ 
         \hline
         ANHIR &  \textbf{0.0384} $\pm$ 0.087  & 0.0588 $\pm$ 0.167 & 0.0333 \\
         \hline
    \end{tabular}
    \label{tab:ablation_table3}
\end{table}
\end{center}

Table \ref{tab:ablation_table3} presents these results. As we can be seen, there is a small drop in performance, but DRMIME still performs better than all the other algorithms with FIRE (Table \ref{tab:fire_res}) and better than most other algorithms with ANHIR (Table \ref{tab:anhir_res}). This comes at a small cost of the optimizer taking longer to converge. It is important to note, that DRMIME results are using only 10\% sampling, whereas the other algorithms use 50\% sampling (see Appendix) due to limited memory available on the GPU.

% \subsection{Effect of Downsampling Factor}
% In all our experiments, we downscale by a factor of 2 in the gaussian pyramid. In order to study the effect of this scaling factor on the algorithm, we tried out factors of 1.5 and 1.25 on the FIRE dataset as well.
% \begin{center}
%     \begin{tabular}{||c|c||}
%     \hline
%     \textbf{Downscale Factor} & \textbf{NAED} \\
%     \hline \hline
%       2  & 0.0048724 $\pm$ 0.0141651 \\
%       \hline
%         1.5 & 0.0044063 $\pm$ 0.0151557\\
%         \hline
%         1.25 & 0.0066397 $\pm$ 0.0198864\\
%     \hline
%     \end{tabular}
% \end{center}

\section{Conclusion and Future Work}
Although here our parametrization limits our ability to affine/perspective transforms, the idea should be extendable to deformable image registration once parametrized appropriately. Also, our experiments were limited to 3 channel RGB images. Since MINE scales linearly with dimensionality, it can be applied to even higher dimensional images. This means that DRMIME could be used for hyperspectral/multispectral image registration as well.

\section{Appendix}
\label{sec:appendix}
\subsection{DV Lower Bound Reaches Mutual Information}
MINE maximizes the DV lower bound (\ref{eqn:DV}) with respect to a function $f(x,z)$. Let us consider a perturbation function $g(x,z)$ and the perturbed objective function $J(f+\epsilon g)$ for a small number $\epsilon$. Taking the following limit (using L\textquotesingle Hospital\textquotesingle s rule), we obtain:
\begin{equation}
\begin{split}
   & \lim_{\epsilon\to0}\frac{J(f+\epsilon g) - J(f)}{\epsilon} = \int g(x,z)P_{XZ}(x,z)dxdz -\\
   &  \int g(x,z)\frac{exp(f(x,z))P_{X}(x)P_{Z}(z)}{\int exp(f(x,z))P_{X}(x)P_{Z}(z)dxdz}]dxdz.
\end{split}
\end{equation}
Using principles of calculus of variations\cite{gelfand2000calculus}, this limit should be 0 for $J$ to achieve an extremum. Since perturbation function $g(x,z)$ is arbitrary, this condition is possible only when
\begin{equation}
P_{XZ}(x,z) = \frac{exp(f(x,z)) P_{X}(x) P_{Z}(z)}{\int exp(f(x,z)) P_{X}(x) P_{Z}(z) dxdz},
\label{eq:gibbs}
\end{equation}
i.e., the Gibbs density \cite{belghazi2018mine} is achieved.
From (\ref{eq:gibbs}), we obtain:
\begin{equation}
\begin{split}
    & f(x,z) = \\
    & log(\frac{P_{XZ}(x,z)}{P_{X}(x)P_{Z}(z)} \int exp(f(x,z)) P_{X}(x) P_{Z}(z) dxdz).
\end{split}
\end{equation}
Using this expression in equation (\ref{eqn:DV}), we obtain:
\begin{equation}
    J(f) = \int P_{XZ}(x,z) log \frac{P_{XZ}(x,z)}{P_{X}(x)P_{Z}(z)} dx dz = MI.
\end{equation}
Thus, maximization of $J(f)$ leads to mutual information.
\subsection{Algorithm Hyperparameters}

All architectures and hyper-parameters for our experiments are listed here:
\subsubsection{DRMIME}:
\begin{enumerate}
    \item learningRate: $\alpha= 1e-3$, $\beta = 5e-3$, $\gamma = 1e-4$
    \item numberOfIterations: 500 (FIRE)/1500 (ANHIR)
    \item Optimizer : ADAM with AMSGRAD
\end{enumerate}

\subsubsection{MMI}:
\begin{enumerate}
    \item learningRate: 1e-5
    \item numberOfIterations: 5000
    \item numberOfHistogramBins: 100
    \item convergenceMinimumValue: 1e-9
    \item convergenceWindowSize: 200
    \item SamplingStrategy: Random
    \item SamplingPercentage: 0.5
\end{enumerate}

\subsubsection{JHMI}:
\begin{enumerate}
    \item learningRate: 1e-1
    \item numberOfIterations: 5000
    \item numberOfHistogramBins: 100
    \item convergenceMinimumValue: 1e-9
    \item convergenceWindowSize: 200
    \item SamplingStrategy: Random
    \item SamplingPercentage: 0.5
\end{enumerate}

\subsubsection{MSE}:
\begin{enumerate}
    \item learningRate: 1e-6
    \item numberOfIterations: 5000
    \item convergenceMinimumValue: 1e-9
    \item convergenceWindowSize: 200
\end{enumerate}

\subsubsection{NCC}:
\begin{enumerate}
    \item learningRate: 1e-1
    \item numberOfIterations: 5000
    \item convergenceMinimumValue: 1e-9
    \item convergenceWindowSize: 200
\end{enumerate}

\subsubsection{NMI}:
\begin{enumerate}
    \item numberOfIterations: 5000
\end{enumerate}

\subsubsection{AMI}:
\begin{enumerate}
    \item learningRate: 1e-4
    \item numberOfIterations: 5000
    \item Optimizer : AMSGRAD
\end{enumerate}

\bibliography{bibfile}
\bibliographystyle{IEEEtran}

\end{document}